\documentclass[sigconf,authorversion,nonacm]{acmart}

\usepackage{dsfont}
\usepackage{caption}
\usepackage{subcaption}
\usepackage{rotating}
\usepackage{arydshln}
\usepackage{mathrsfs}

\newtheorem{definition}{Definition}

\AtBeginDocument{%
  \providecommand\BibTeX{{%
    \normalfont B\kern-0.5em{\scshape i\kern-0.25em b}\kern-0.8em\TeX}}}
 
\begin{document}

\title{Anomaly Detection Based on Isolation Mechanisms: A Survey}

\author{Yang Cao}
\affiliation{%
  \institution{Centre for Cyber Resilience and Trust\\ Deakin University} 
  \city{Burwood, VIC}
  \country{Australia}}
\email{charles.cao@ieee.org }

\author{Haolong Xiang}
\affiliation{%
  \institution{School of Computing\\
    Macquarie University} 
  \city{Macquarie Park, NSW}
  \country{Australia}}
  \email{haolong.xiang@mq.edu.au}

 \author{Hang Zhang} 
 \affiliation{
 	\institution{State Key Laboratory
of Novel Software Technology\\ Nanjing University}
\city{Jiangsu} 
 \country{China} 
 }
 \email{zhanghang@lamda.nju.edu.cn}

\author{Ye Zhu}
\authornote{Corresponding Author}
\affiliation{%
  \institution{Centre for Cyber Resilience and Trust\\ Deakin University}
  \city{Burwood, VIC}
  \country{Australia}}
\email{ye.zhu@ieee.org }

 \author{Kai Ming Ting} 
 \affiliation{
 	\institution{State Key Laboratory
of Novel Software Technology\\ Nanjing University}
\city{Jiangsu} 
 \country{China} 
 }
 \email{tingkm@nju.edu.cn}


\begin{abstract}
Anomaly detection is a longstanding and active research area that has many applications in domains such as finance, security, and manufacturing. However, the efficiency and performance of anomaly detection algorithms are challenged by the large-scale, high-dimensional, and heterogeneous data that are prevalent in the era of big data. Isolation-based unsupervised anomaly detection is a novel and effective approach for identifying anomalies in data. It relies on the idea that anomalies are few and different from normal instances, and thus can be easily isolated by random partitioning. Isolation-based methods have several advantages over existing methods, such as low computational complexity, low memory usage, high scalability, robustness to noise and irrelevant features, and no need for prior knowledge or heavy parameter tuning. In this survey, we review the state-of-the-art isolation-based anomaly detection methods, including their data partitioning strategies,  anomaly score functions, and algorithmic details.  We also discuss some extensions and applications of isolation-based methods in different scenarios, such as detecting anomalies in streaming data, time series, trajectory, and image datasets. Finally, we identify some open challenges and future directions for isolation-based anomaly detection research.
\end{abstract}



\maketitle

\section{Introduction}

Anomaly detection refers to the identification of data points that significantly differ from the normal patterns in a dataset~\cite{chandola2009anomaly,pang2021deep}. The study of anomalies in data dates back to the 19th century in statistics~\cite{edgeworth1887xli}. In recent years, anomaly detection has been a crucial task in various research and industrial domains, such as health care~\cite{fernando2021deep,venkataramanaiah2020ecg},  finance~\cite{huang2022dgraph,kumar2021machine}, security~\cite{cui2021security,hosseinzadeh2021improving}, energy~\cite{erhan2021smart,finke2021autoencoders} and manufacturing~\cite{alfeo2020using,kamat2020anomaly,pittino2020automatic}, where it can help monitor, control, and secure various processes and systems.



Compared to other types of anomaly detection methods, isolation-based unsupervised methods have several advantages, such as linear time complexity, small memory requirement, unsupervised learning, robustness to noise, irrelevant attributes, and data density distribution, making them a valuable tool for both researchers and practitioners. The main idea of isolation-based methods is to employ an isolation mechanism to construct isolating partitions in the input data space, such that anomalies are more likely to be isolated from the rest of the data. The first and most representative isolation-based method is Isolation Forest (iForest)~\cite{liu2008isolation}. Building upon the success of iForest, many extensions and variations have been proposed to address different challenges and improve performance, such as iNNE~\cite{bandaragoda2018isolation}, Extended Isolation Forest~\cite{hariri2019extended}, OptiForest~\cite{xiang2023optiforest}, and Isolation Distributional Kernel~\cite{ting2021isolation,liu2012isolation,wang2024detecting}. These methods usually identify various kinds of anomalies with no learning. 

This paper is the first comprehensive survey of isolation-based unsupervised anomaly detection. It reviews the anomaly detection methods based on isolation mechanisms, covering their data partitioning, anomaly scoring, and algorithmic aspects. It also explores some extensions and applications of isolation-based methods for various scenarios, such as streaming data, time series, trajectory, and image datasets. Lastly, it highlights some open challenges and future directions.



\section{Anomaly definition}~\label{sec: definition}

\begin{figure}
     \centering
     \begin{subfigure}[b]{0.23\textwidth}
         \centering
         \includegraphics[width=.8\textwidth]{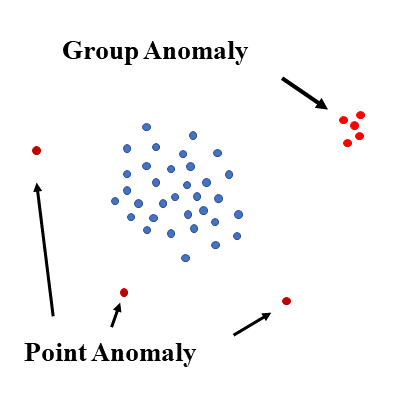}
         \caption{Point \& Group anomaly}
         \label{fig:point}
     \end{subfigure}
     \begin{subfigure}[b]{0.23\textwidth}
         \centering
         \includegraphics[width=.8\textwidth]{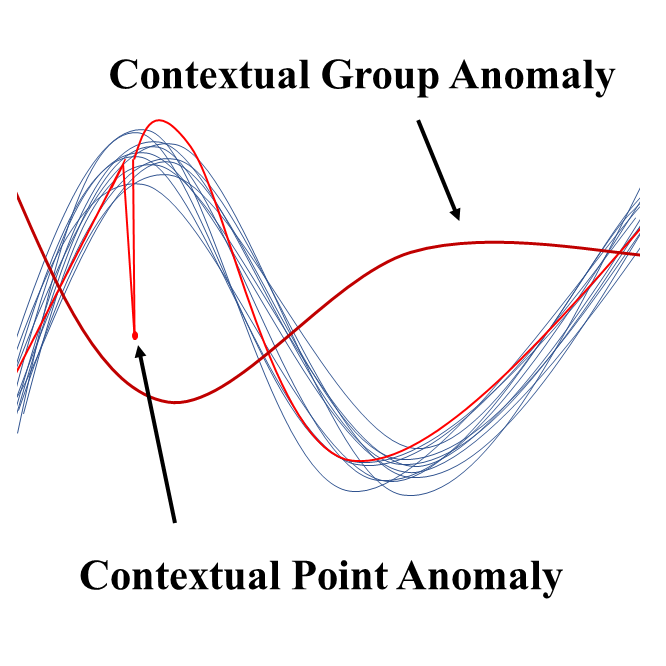}
         \caption{Contextual anomaly}
         \label{fig:contextual}
     \end{subfigure}
     \begin{subfigure}[b]{0.23\textwidth}
         \centering
         \includegraphics[width=.8\textwidth]{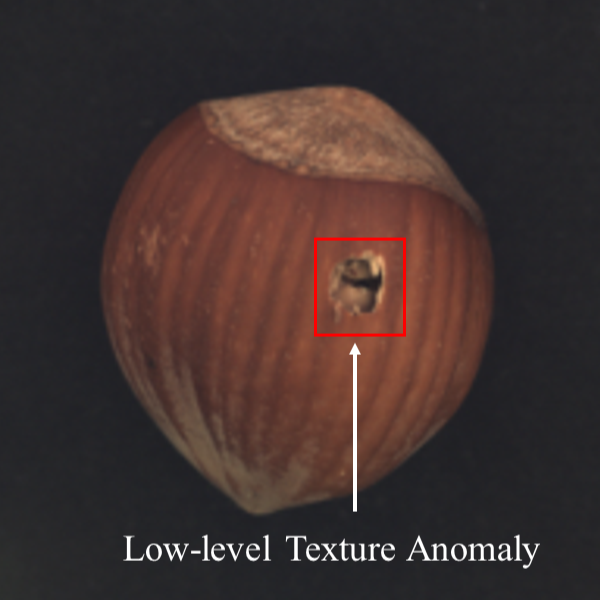}
         \caption{Low-level texture anomaly}
         \label{fig:low-level}
     \end{subfigure}
     \begin{subfigure}[b]{0.23\textwidth}
         \centering
         \includegraphics[width=0.8\textwidth]{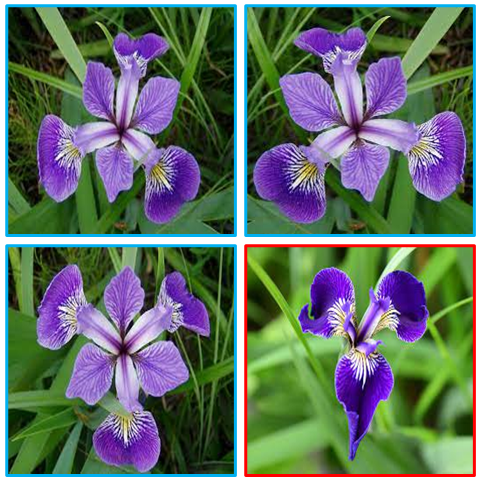}
         \caption{High-level semantic anomaly}
         \label{fig:high-level}
     \end{subfigure}
        \caption{Demonstration of different types of anomalies. (a) A point anomaly is a single abnormal red point; a group anomaly is an abnormal red cluster. (b) A contextual group anomaly is a series of related but individually normal data points that are collectively unusual. (c) A breakage on the texture of a nut (d) Four iris images, three are Versicolour and one is Setosa marked in red, they have similar texture in low-level pixel but with different shapes~\cite{misc_iris_53}.}
        \label{fig:types}
\end{figure}

An anomaly is an observation that significantly deviates from some concept of normality~\cite{hawkins1980identification}. The existing anomaly types can be broadly categorized as follows:
\begin{itemize}
    \item \emph{Point anomalies} are individual instances that differ markedly from the majority of the data~\cite{chandola2009anomaly}.
    \item \emph{Group or Collective anomalies} anomalies are a subset of data points that exhibit anomalous behavior as a whole compared to the rest of the data~\cite{song2007conditional, gupta2013outlier, chandola2010anomaly}.
    \item \emph{Contextual or Conditional anomalies} anomalies are data points or groups that are only anomalous in a specific context or condition~\cite{chandola2009anomaly, ruff2021unifying}.
\end{itemize}

For unstructured data such as images and texts, anomalies can be distinguished at two levels as follows~\cite{ahmed2020detecting, ruff2021unifying}: 
\begin{itemize} 
    \item \emph{Low-level sensory anomalies} are data instances that differ from the expected low-level features, which correspond to pixel-level characteristics, such as texture defects or artifacts in images.
    \item \emph{High-level semantic anomalies} are data instances that violate the expected high-level features, which correspond to semantic concepts or topics in the text, such as the class of the object.
\end{itemize}

Figure~\ref{fig:types} shows the 6 kinds of anomalies on different datasets.

\section{Isolation partitioning strategies}~\label{sec:partitioning}

In this section, we introduce five main space partitioning methods used in isolation-based anomaly detection methods. Table~\ref{tab:part} shows examples of different partitioning strategies on a two-dimensional dataset.

\begin{table}[t]
    \centering
    \caption{Isolation partitioning strategies. The partitionings are constructed using the subsample points marked in red, and the numbers shown in the cells are the index of the corresponding cell. The feature maps of x and y are given below the figure.}
    \label{tab:part}
    \begin{tabular}{ccc}
    \toprule
        Partitioning  &   {Demo 1}   &   {Demo 2}  \\
    \midrule
        \begin{turn}{90}    Axis parallel \end{turn}  & \includegraphics[width=1.2in]{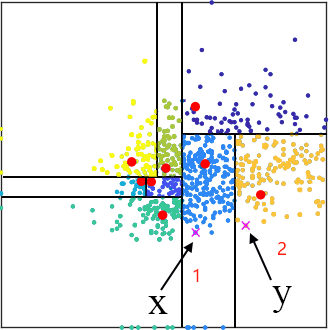} &\includegraphics[width=1.2in]{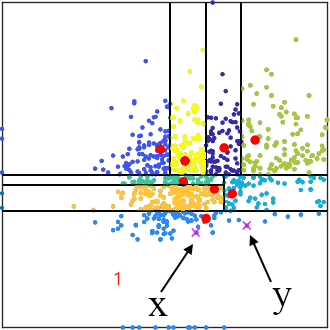}  \\
        & $\Phi(x) = [1~0~0~0~0~0~0~0]$ & $\Phi(x)  = [1~0~0~0~0~0~0~0]$\\
        & $\Phi(y)  = [0~1~0~0~0~0~0~0]$ & $\Phi(y) = [1~0~0~0~0~0~0~0]$\\ \midrule
        \begin{turn}{90}   Random hyperplane \end{turn} & \includegraphics[width=1.2in]{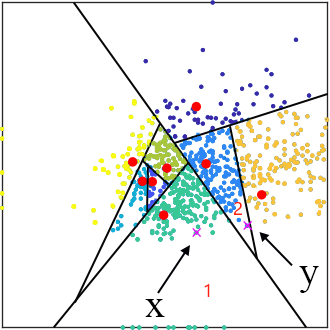} & \includegraphics[width=1.2in]{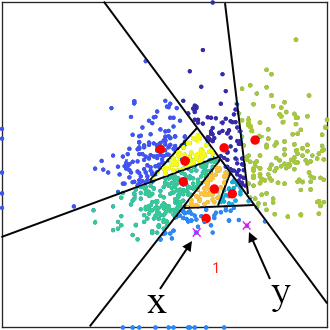}   \\
        & $\Phi(x)  = [1~0~0~0~0~0~0~0]$ & $\Phi(x)  = [1~0~0~0~0~0~0~0]$\\
        & $\Phi(y)  = [0~1~0~0~0~0~0~0]$ & $\Phi(y) = [1~0~0~0~0~0~0~0]$\\ \midrule
        \begin{turn}{90}  Hypersphere \end{turn}  & \includegraphics[width=1.2in]{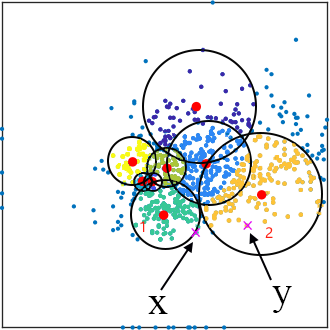} &\includegraphics[width=1.2in]{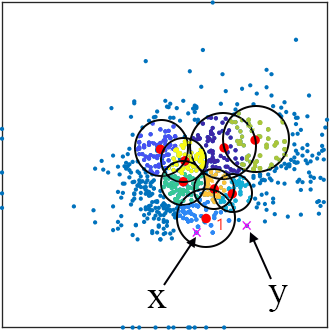}     \\
        & $\Phi(x)  = [1~0~0~0~0~0~0~0]$ & $\Phi(x)  = [1~0~0~0~0~0~0~0]$\\
        & $\Phi(y)  = [0~1~0~0~0~0~0~0]$ & $\Phi(y) = [0~0~0~0~0~0~0~0]$\\ \midrule
        \begin{turn}{90}   Vonoroi diagram   \end{turn}   &\includegraphics[width=1.2in]{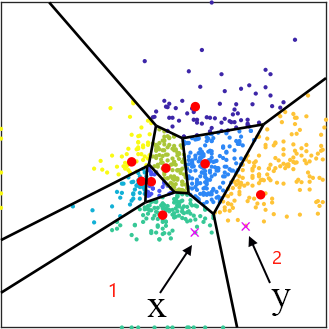} & \includegraphics[width=1.2in]{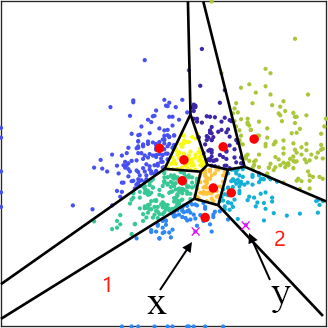}   \\
        & $\Phi(x)  = [1~0~0~0~0~0~0~0]$ & $\Phi(x)  = [1~0~0~0~0~0~0~0]$\\
        & $\Phi(y)  = [0~1~0~0~0~0~0~0]$ & $\Phi(y) = [0~1~0~0~0~0~0~0]$\\
    \bottomrule
    \end{tabular}
\end{table}

\subsection{Axis parallel splitting}

Axis parallel splitting is a common method for constructing tree-based models by dividing the feature space into rectangles (or hyper-rectangles in higher dimensions), with each leaf node in the tree representing one of these rectangles. An important difference in data splitting between random forest and isolation forest is that random forest chooses a feature and a threshold that maximizes the information gain or minimizes the Gini impurity, which depends on the class labels of the data points. While isolation forest randomly selects a feature, followed by the random determination of a split value within the range of the chosen feature.

\subsection{Non-axis parallel splitting}

The axis-parallel splitting of data space can cause a specific limitation, i.e., it obscures anomalies that align with the axes of normal data clusters. \cite{bandaragoda2018isolation,laskar2021extending}. To overcome this issue, many non-axis parallel splitting strategies have been proposed.

\subsubsection{Random hyperplane}

SCiForest uses random hyperplanes to replace axis parallel splitting to partition the data space in isolation forest~\cite{liu2010detecting}. Random hyperplanes are similar to oblique hyperplanes which have been studied in the decision tree, both of them are non-axis parallel to the original feature~\cite{murthy1994system}. The difference between them is that the attributes and coefficients are chosen randomly in SCiForest, but they are chosen to minimize the impurity based on labels in an oblique decision tree. 

\subsubsection{Hypersphere}

iNNE~\cite{bandaragoda2018isolation} constructs hyperspheres that enclose $x$ and no other points in the training set to isolate each point $x$. The radius of a hypersphere is defined by the distance from  $x$ to its nearest neighbor within the training set. The hypersphere size reflects the local density: sparse regions have large hyperspheres and dense regions have small hyperspheres. Since the anomalies are more likely to be in the sparse regions, the hypersphere size with larger values indicates a higher anomaly likelihood. The hypersphere size is analogous to the isolation tree approach, while isolation trees assign shorter paths to anomalies, iNNE assigns larger hyperspheres to anomalies.

\subsubsection{Voronoi diagram}

Axis-parallel splitting often results in overextended partitions near the root of the unbalanced tree, as shown by the elongated rectangles ~\cite{qin2019nearest}. To avoid this problem, aNNE~\cite{qin2019nearest} employs the nearest neighbor mechanism to create a Voronoi diagram~\cite{aurenhammer1991voronoi} for data space partitioning. A Voronoi diagram divides the plane into regions, called Voronoi cells (isolated partition), based on the proximity to a set of points. Each point has a Voronoi cell that contains all the points closer to it than to any other point. The Voronoi cell boundary is the perpendicular bisector of the line segment between the points. 
Note that Voronoi cells satisfy the criteria of large partitions in sparse regions and small partitions in dense regions, without producing the negative effect of elongated rectangles.

\subsubsection{Hash-based splitting}

LSHiForest~\cite{zhang2017lshiforest} extends the binary tree of isolation forest to a multi-fork tree by using locality-sensitive hashing (LSH), which hashes similar points into the same bins with high probability.

\begin{figure}[!tb]
     \centering
     \begin{subfigure}[b]{0.23\textwidth}
         \centering
         \includegraphics[width=0.85\textwidth]{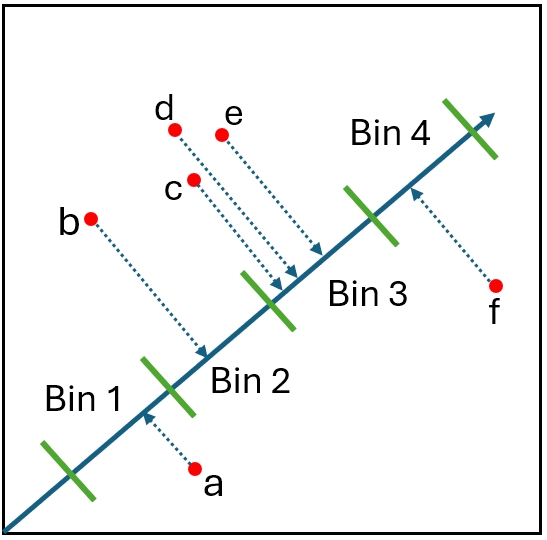}
         \caption{Data projection and hash}
         \label{fig:T1}
     \end{subfigure}
     \begin{subfigure}[b]{0.23\textwidth}
         \centering
         \includegraphics[width=.65\textwidth]{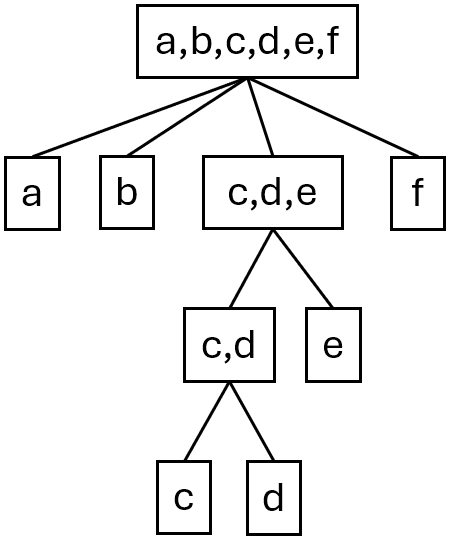}
         \caption{One LSHiTree}
         \label{fig:T2}
     \end{subfigure}
        \caption{Illustration of LSHiForest.}
        \label{fig:T}
\end{figure}

Given a node $T$, the main steps of hash-based splitting include (1) projecting each point $x$ from the node on a random vector by using  $p_{a,b}(x)=ax + b$ where $a$ is a random vector with components drawn from the $p$-stable distribution independently, $b$ is an offset value uniformly sampled from $[0,W]$ and $W$ is a predefined parameter as the hashing bin size; (2) dividing the projected values into $W$ equal-sized sequential intervals stating from the first projected value, each interval can be treated as an LSH bin, i.e., $f_{a,b}(x)= \lfloor \frac{p_{a,b}(x)}{W} \rfloor$. Then, each bin becomes a branch and the node $T$ will have $W$ branches. The branch node with more than one point is further divided using the same strategy, but the projection is still based on the full original data attribute.

Figure \ref{fig:T} illustrates the projection process of a single LSHiTree on 6 data points. Each level of the tree may have a different number of branches, thus, reducing the height of the tree for higher efficiency in anomaly score calculation.

\section{Point anomaly detection}~\label{sec:point}
This section delineates the computational methodologies employed to ascertain the anomaly score of a data point, which is predicated on the geometric characteristics inherent to the partitioning process. Subsequently, it expounds upon the applications of point anomaly detection that utilize isolation-based approaches.

\subsection{Geometric-based perspective}

\begin{figure*}
     \centering
     \begin{subfigure}[b]{0.44\textwidth}
         \centering
         \includegraphics[width=.93\textwidth]{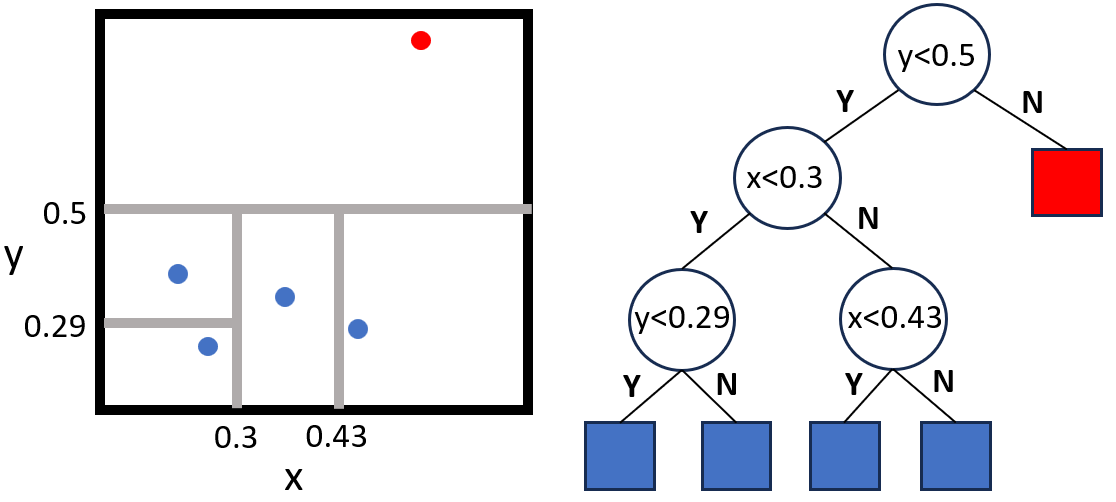}
         \caption{Isolation tree partitioning}
         \label{fig:ifsplit}
     \end{subfigure}
     \begin{subfigure}[b]{0.44\textwidth}
         \centering
         \includegraphics[width=.73\textwidth]{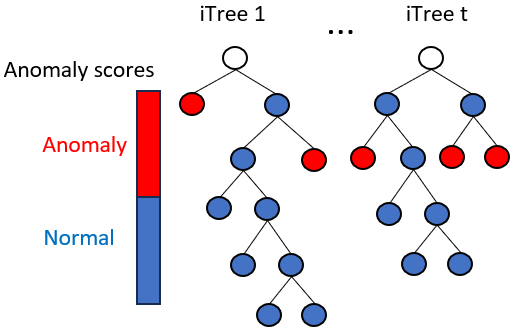}
         \caption{Isolation trees}
         \label{fig:itrees}
     \end{subfigure}
        \caption{Illustration of isolation forest.}
        \label{fig:iforest}
\end{figure*}

\subsubsection{Path length}

Since recursive partitioning can be represented by a tree structure, the number of partitions required to isolate a point is equivalent to the path length from the root node to a terminating node~\cite{liu2008isolation}. 
An Isolation Tree (iTree) constitutes a binary tree wherein each non-terminal node bifurcates into precisely two child nodes.

Given a dataset with $n$ points, assuming all points are distinct, each point is isolated to an external node when an iTree is fully grown, in which case the number of external nodes is $n$ and the number of internal nodes is $n-1$; the total number of nodes of an iTrees is $2n-1$. Path Length $h(x)$ of a point $x$ is measured by the number of edges $x$ that traverses an iTree from the root node until the traversal is terminated at an external node.

Because iTrees have the same structure as Binary Search Trees (BSTs),  the estimation of average $h(x)$ for external node terminations is the same as the unsuccessful search in BST. The average of $h(x)$ can be estimated as~\cite{liu2008isolation}:

\begin{equation}
    c(n) = 2H(n-1)-(2(n-1)/n),
\end{equation}
where $H(i)$ is the harmonic number and it can be estimated by $ln(i) + 0.577$ (Euler’s constant). $c(n)$ is the average of $h(x)$ given $n$ and is used to normalise $h(x)$. The anomaly score $s$ of an instance $x$ is defined as:
\begin{equation}
    s(x, n) = 2^{-\frac{E(h(x))}{c(n)}}, 
\end{equation}
where $E(h(x))$ is the average of $h(x)$ from a collection of isolation trees.

Figure~\ref{fig:iforest} demonstrates the partitioning result and ensemble process of isolation forest.

\subsubsection{Hypersphere size}

iNNE constructs hyperspheres to isolate each instance in a subsample and builds an ensemble from multiple subsamples. 
Let $\mathcal{D} \subset D$ be a subsample of size $\psi$ selected randomly without replacement from a data set $D \subset \mathbb{R}^d$, and let $\eta_x$ be the nearest neighbors of $x \in \mathbb{R}^d$. A hypersphere $B(c)$ centered at $c$ with radius $\tau(c) = ||c -\eta_c||$ is defined to be $\{x:||x-c|| < \tau(c)\}$, where $c, \eta_c \in \mathcal{D}$ and $||\cdot||$ is Euclidean distance measure.

It should mentioned that $B(c)$ is the maximal hypersphere, which isolates instance $c$ from all other instances in $\mathcal{D}$. The radius $\tau(c)$ quantifies how isolated $c$ is, a larger $\tau(c)$ implies greater isolation of $c$, and vice versa. Rather than a global measure, iNNE uses a local measure of isolation, which is the ratio of $B(c)$ and $B(\eta_c)$, a measure of isolation of $c$ relative to its neighborhood is defined as: 

\begin{equation}
I(x) = \left\{
\begin{array}{l l}
   1 - \frac{\tau(\eta_{cnn(x)})}{\tau(cnn(x))}, & \text{if } x \in \bigcup\limits_{c \in  \mathcal{D}}B(c) \\
    1, & \text{otherwise}
  \end{array} \right.
\end{equation}
where $cnn(x) = {\arg\min}_{c\in \mathcal{D}} \{\tau_c:x\in B(c) \}$.

$I(x)$ takes values in the range of $[0, 1]$, because $\frac{\tau(\eta_{cnn(x)})}{\tau(cnn(x))}\leq1$. 
If $x$ lies outside all hyperspheres, $x$ is assumed to be distant from all instances in $\mathcal{D}$ and is given the highest isolation score.
Since iNNE has $t$ sets of hyperspheres, the anomaly score of $x$ can be defined as:
\begin{equation}
    \bar{I}(x) = \frac{1}{t} \sum_{i=1}^{t}I_i(x)
\end{equation}
where $I_i(x)$ is the isolation score based on $\mathcal{D}_i$.

\begin{figure}
    \centering
    \includegraphics[width = 0.24\textwidth]{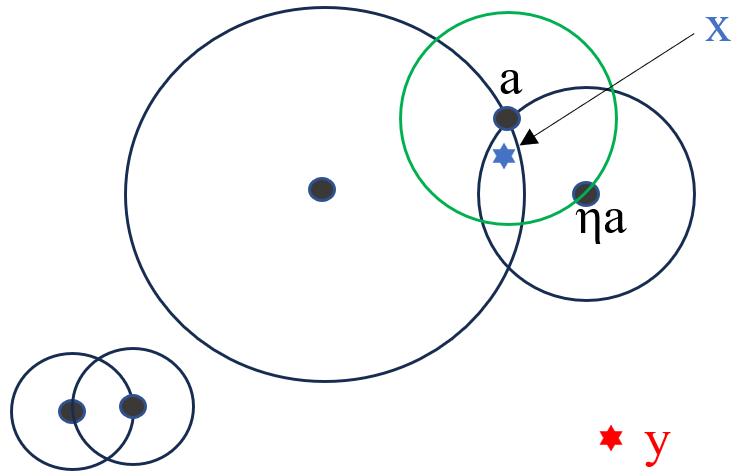}
    \caption{Illustration of iNNE with $\psi$ = 5. If $x$ is in the overlap region of multiple partitions, it will be assigned to the hypersphere generated by $x$'s nearest subsample point. Since $y$ is out of any hypersphere, the anomaly score will be 1.}
    \label{fig:inne}
\end{figure}

Figure~\ref{fig:inne} shows the isolation scores determined for $x$ and $y$ using the isolation model.

\subsection{Similarity-based perspective}
Anomaly detection can be performed by using a similarity measure, which quantifies how similar different points in a dataset are. According to this approach, the anomalies are the points that have a low similarity to most of the other points, which are considered normal. Based on the partitioning, there are two methods to calculate the similarity between points, including mass-based and kernel-based.

\subsubsection{Mass-based}
Let $D \subset \mathcal{X} \subseteq \mathbb{R}^d$ be a dataset sampled from an unknown probability distribution $\mathcal{P}_D$; 
Each $H$ denotes a partitioning from each $\mathcal D$, let $\mathds{H}_\psi(D)$ consist of all partitioning $H$ admissible from the given dataset $\mathcal{D} \subset D$,
where each point $z \in \mathcal{D}$ has the same probability of being selected from $D$.
Each of the $\psi$ isolating partitions $\theta[{\mathbf{z}}] \in H$ isolates a data point ${ \mathbf{z}}$ from the rest of the points in a random subset $\mathcal D$, and $|\mathcal D|=\psi$. 

\begin{definition}
    $R(x, y| H; D)$ is the smallest local region covering both $x$ and $y$ w.r.t $H$ and $D$ is defined as:
    \begin{equation}
        R(x, y| H; D)=\mathop{\arg\min}\limits_{\theta[{\mathbf{z}}] \in H, s.t.\{x,y\}\in \theta[{\mathbf{z}}]} \sum_{l \in D}\mathds{1}(l \in \theta[{\mathbf{z}}])
    \end{equation}
    where $\mathds{1}$ is an indicator function, $l$ is the data points in dataset $D$; Note that $R(x, y\mid H; D)$ is the smallest local region covering $x$ and $y$, it is analogous to the shortest distance between $x$ and $y$ used in the geometric model. 
\end{definition}

\begin{definition}
    Mass-based dissimilarity of $x$ and $y$ w.r.t $D$ and $F$ is defined as the expected probability of $R(x, y \mid H; D)$:
    \begin{equation}
        M(x, y \mid D, F) = E_{\mathds{H}_\psi(D)}[\mathcal{P}_F(R(x, y| H; D))]
    \end{equation}
    where $\mathcal{P}_F$ is the probability w.r.t $F$; and the expectation is taken over all models in $\mathds{H}_\psi(D)$.
\end{definition}

In practice, the dissimilarity can be estimated as follows:
\begin{equation}
    M_E(x, y \mid D) = \frac{1}{t} \sum_{i=1}^{t}\mathcal{\tilde{P}}(R(x, y \mid H_i; D))
\end{equation}
where $\mathcal{\tilde{P}} = \frac{1}{|D|} \sum_{l\in D} \mathds{1}(l\in R)$.

\subsubsection{Kernel-based}

The main idea of the isolation kernel is to use a data space partitioning strategy to estimate how likely two points will be partitioned into the same partition. The unique characteristic of Isolation Kernel is that ``\textbf{two points in a sparse region are more similar than two points of equal inter-point distance in a dense region}'' \cite{qin2019nearest, ting2018isolation}. Furthermore, Isolation kernel has a finite number of features that enable fast feature mapping and similarity calculation. With these two characteristics, both global and local anomalies can be detected efficiently.\footnote{A data point is an anomaly if it differs significantly from either its local neighborhood (local anomaly) or the whole data set (global anomaly). Local anomalies tend to have a lower density than their surrounding region \cite{breunig2000lof}.}.

\begin{definition}\label{def:def1}
	\textbf{Isolation Kernel~\cite{qin2019nearest, ting2018isolation}}. Given any two points $x,y \in \mathbb{R}^d$, 	the isolation kernel is defined as the expectation taken over the probability distribution on all partitionings $H \in \mathbb{H}_\psi(D)$, that both $x$ and $y$ all fall into the same isolation partition $\theta[z] \in H$	with the condition $z \in \mathcal{D} \subset D$ and $\psi = |\mathcal{D}|$:
	\begin{eqnarray}
		\kappa(x,y\mid H; D)& = &\mathbb{E}_{\mathbb{H}_\psi(D)}[\mathds{1}(x,y \in \theta[z] \mid \theta [z] \in H)] \nonumber \\
		& = &\frac{1}{t}\sum_{i=1}^{t} \mathds{1}(x,y \in \theta \mid \theta \in H_i) \nonumber \\
		& = &\frac{1}{t}\sum_{i=1}^{t}\sum_{\theta \in H_i}\mathds{1}(x\in \theta)\mathds{1}(y\in \theta),
		\label{equ:ik}
	\end{eqnarray}
	where $\mathds{1}$ is an indicator function; $\kappa$ is the Isolation kernel function which is constructed based on a finite number of partitionings $H_i, i=1,...,t$	and each $H_i$ is generated using $\mathcal{D}_i \subset D$; and	$\theta$ is a shorthand $\theta[z]$.
\end{definition}

\begin{definition}\label{def:def2}
	\textbf{Feature map of Isolation Kernel~\cite{ting2021isolation}}.
	For point $x \in \mathbb{R}^d$,
	the feature map $\Phi:
		x \rightarrow \{0,1\}^{t\times \psi}$
	of Isolation kernel $\kappa$ is a binary vector
	which indicates the partitionings $H_i \in \mathbb{H}_\psi(D)$,
	$i=1, \cdot,t$;
	where the point $x$ falls into only one of the $\psi$ partitions or none in each partitioning $H_i$.
\end{definition}

Figure \ref{tab:part} demonstrates the feature map with different partitioning strategies when $\psi=8$ and $t=2$. For example, when using hypersphere for partitioning twice as shown in Figure \ref{tab:part}, the feature map of $x$ is the concatenation of the two binary vectors as $\Phi(x)=[1\ 0\ 0\ 0\ 0\ 0\ 0\ 0\ 1\ 0\ 0\ 0\ 0\ 0\ 0\ 0]$.

Based on the feature map,
we can re-write Equation~\ref{equ:ik} as using $\Phi$ gives:
\begin{equation}\label{equ:ec4}
	\kappa(x,y\mid H; D)=\frac{1}{t}\left\langle\Phi(x\mid D),\Phi(y\mid D)\right\rangle
\end{equation}

As demonstrated in Figure \ref{tab:part}, when using hypersphere for partitioning, the similarity $\kappa(x,y\mid H; D)=0$. However, $\kappa(x,y\mid H; D)=0.5$ if using random projection.  

Once we get the feature map of all points or obtain the pairwise similarities between all points, we can apply any existing point anomaly detectors, depending on their input requirements.

\subsection{Other applications}

The above mentioned methods can be directly used in many academic and industry fields with static, numerical and structured data, such as sensors~\cite{du2020itrust, wang2019isolation, fang2023detecting}, network security~\cite{marteau2021random, chiba2019newest}, financial~\cite{hilal2022financial}, medical~\cite{islah2020machine, guo2022epileptic, zhao2024outlier} and energy~\cite{himeur2021artificial, ahmed2019unsupervised}. They can also be applied to other types of datasets as follows.

\subsubsection{Streaming data}
Streaming data are independent and identically distributed (i.i.d.) and have several distinctive features, such as being real-time, continuous, ordered, large data volume, and infinity~\cite{goldenberg2019survey, gama2014survey}. Most streaming anomaly detection algorithms are based on static anomaly detectors, but with additional mechanisms to handle the dynamic nature of streaming data. These algorithms usually consist of two components: an anomaly detector that assigns anomaly scores to the incoming data points, and a model updater that adapts the underlying model to the changing data characteristics. 

\begin{figure}[!tb]
    \centering
    \includegraphics[width=.4 \textwidth]{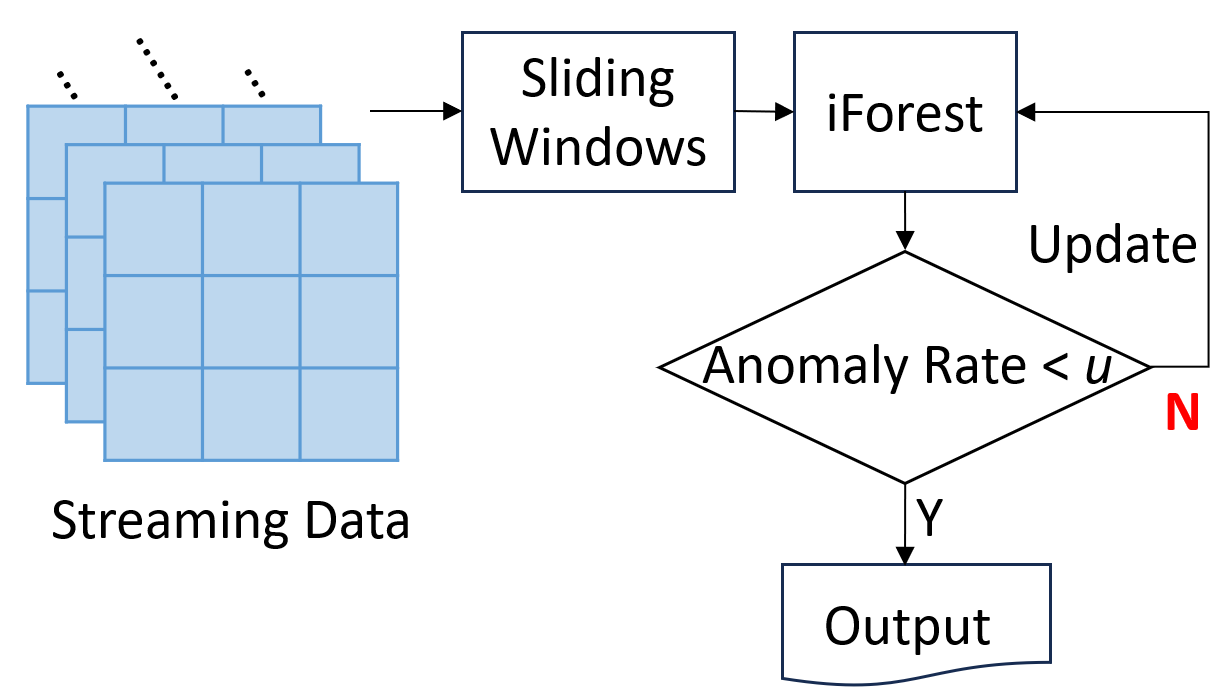}
    \caption{Illustration of the general procedure of iForestASD}
    \label{fig:iforestasd}
\end{figure}

Isolation-based methods have been commonly used as the static anomaly detector in some streaming anomaly algorithms, e.g., IDForest~\cite{xiang2022edge}, iForestASD~\cite{ding2013anomaly}, HS-Trees~\cite{tan2011fast}, RRCF~\cite{guha2016robust}, etc. A common procedure used in streaming anomaly detection algorithms: (1) an initial model is trained on some data; (2) new data points are scored as they arrive; (3) since streaming data may exhibit concept drift \cite{gama2014survey}, which refers to the change in data distribution over time, a concept drift detector is also employed in the algorithm. Depending on the concept drift detection, the model is updated with different strategies or remains unchanged.  Figure~\ref{fig:iforestasd} illustrates the procedure of iForestASD~\cite{ding2013anomaly}, which will update the model once the anomaly rate is larger than a threshold.

\begin{figure}[!tb]
    \centering
    \includegraphics[width=.43\textwidth]{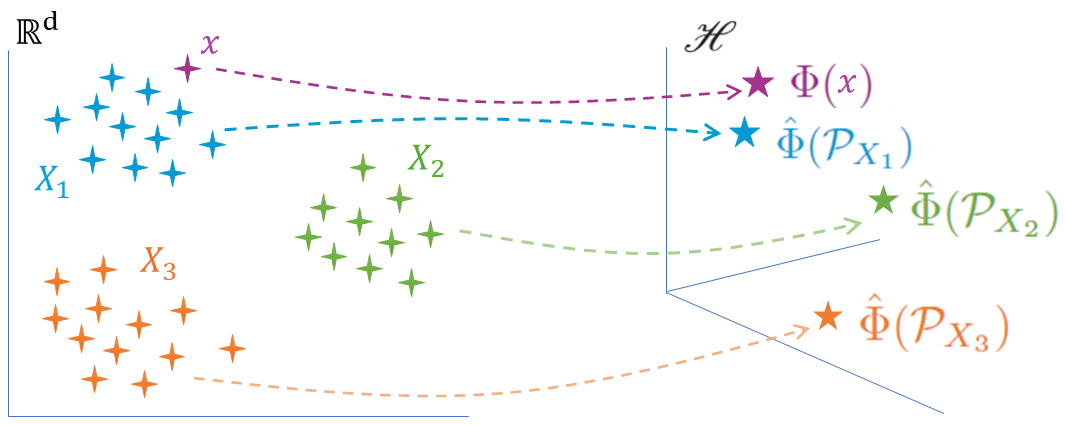}
    \caption{Illustration of kernel mean embedding.}
    \label{fig:ebd}
\end{figure}

\subsubsection{Unstructured data}

Deep neural networks (DNNs) have been widely used in different anomaly detection tasks, particularly on complex data types~\cite{zong2018deep}. By combining DNNs and isolation-based methods, these approaches have extended the applicability of isolation-based methods to unstructured data, e.g. image~\cite{wang2022image, utkin2022improved, zhao2023infrared, li2020out} and hyperspectral~\cite{cheng2023two, song2021spectral, wang2020multiple}, Video~\cite{bhatt2022explosive} and text~\cite{farzad2020unsupervised}.   

Deep isolation forest (DIF)~\cite{xu2023deep} is a representative algorithm that combines DNNs and isolation forest, the general procedures of DIF are (1) map original data into random representation ensembles by using casually initialized neural networks; (2) applies isolation forest to detect anomalies on the extracted representations. By enabling flexible partitioning of the original data space (corresponding to non-linear partitioning on subspaces with different dimensions), this representation scheme of DIF fosters a novel interplay between random representations and isolation based on random partitioning.

\section{Group anomaly detection}~\label{sec:group}

The Isolation Distributional Kernel (IDK)~\cite{ting2021isolation} extends the isolation kernel into a distributional kernel to detect group anomalies by measuring the similarity between distributions, based on Kernel Mean Embedding (KME)~\cite{muandet2017kernel, smola2007hilbert}.

\begin{definition}\label{def:def3}
	\textbf{Isolation Distributional Kernel~\cite{ting2021isolation}}.
	Isolation distributional kernel of two distributions
	$\mathcal{P}_X$ and $\mathcal{P}_Y$
	is defined as:
	\begin{eqnarray}~\label{equ:idk}
 \widehat{\mathcal{K}}(\mathcal{P}_X,\mathcal{P}_Y\mid H; D)
&=& 	\frac{1}{|X||Y|}\sum_{x\in X}\sum_{y\in Y}	\kappa(x,y \mid H; D)\\
  &=&\frac{1}{t|X||Y|}\sum_{x\in X}\sum_{y\in Y}\left\langle\Phi(x\mid D),\Phi(y\mid D)\right\rangle \nonumber  \\
		&=&
 \frac{1}{t}\left\langle\widehat{\Phi}(\mathcal{P}_X|D),\widehat{\Phi}(\mathcal{P}_Y|D)\right\rangle
	\end{eqnarray}
	where $\widehat{\Phi}(\mathcal{P}_X|D) = \frac{1}{|X|}\sum_{x\in X}\Phi(x\mid D)$ is the empirical feature map of kernel mean embedding.
\end{definition}

The similarity score can be normalized to $[0, 1]$ as follows:

\begin{equation} 
	\widehat{\mathcal{K}}(\mathcal{P}_X,\mathcal{P}_Y\ |\ D)
	 =  \frac{\langle  \widehat{\Phi}(\mathcal{P}_X|D), \widehat{\Phi}(\mathcal{P}_Y|D) \rangle}{\sqrt{\langle  \widehat{\Phi}(\mathcal{P}_X|D), \widehat{\Phi}(\mathcal{P}_X|D) \rangle}\sqrt{\langle \widehat{\Phi}(\mathcal{P}_Y|D), \widehat{\Phi}(\mathcal{P}_Y|D) \rangle}}.
	\label{E5}
\end{equation}

The dissimilarity score can be expressed as a distance function $dist_{I}(X,Y)=1 -\widehat{\mathcal{K}}(\mathcal{P}_X,\mathcal{P}_Y\ |\ D)$. It is worth mentioning that IDK can guarantee the `uniqueness' property of this distance function, i.e., $dist_{I}(X,Y)=0$ if and only if $\mathcal{P}_X = \mathcal{P}_Y$.




After obtaining the kernel mean embedding for each distribution, an existing point anomaly detector can be applied to those embedding features, i.e., the identified anomalous points in the embedding space are the group anomalies in the original space.

Figure \ref{fig:ebd} illustrates an example of kernel mean embedding for each of the three groups (sets) $X_1$, $X_2$, and $X_3$. 

Furthermore, let $\Pi=\{g_i | i=1, ..., n\}$ and $g_i=\widehat{\Phi}(\mathcal{P}_{X_i} | D)$,  the feature map of level-2 IDK is: $\widehat{\Phi}_2(\mathcal{P}_\Pi) = \frac{1}{|\Pi|} \sum_{g \in \Pi} \widehat{\Phi}_2(g|\Pi)$. Then we can  use level-2 IDK to calculate the anomaly score\footnote{Kernel functions measure the similarity between data points, so we can use the inverse of the similarity score as an anomaly score, i.e., the higher the similarity score, the more anomalous the data point is.} for each group $X_i$ as follows. 

\begin{equation}
 \widehat{\mathcal{K}}_2(\delta(g_i),\mathcal{P}_\Pi)=\frac{1}{t}\left< \widehat{\Phi}_2(\delta(g_i)), \widehat{\Phi}_2(\mathcal{P}_\Pi) \right>,
 \label{IDK2}
\end{equation}


\noindent where $\delta(g_i)$ is a Dirac measure of point $g_i$.

\begin{figure}[!tb]
    \centering
    \includegraphics[width=.49\textwidth]{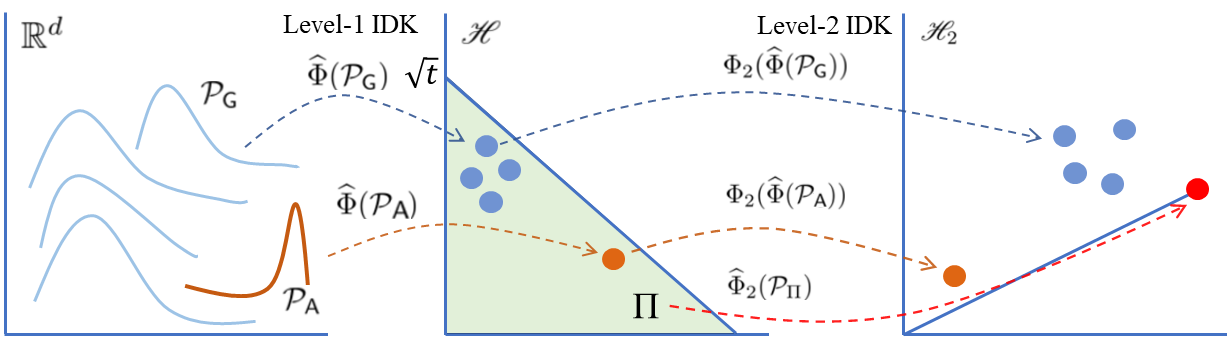}
    \caption{Illustration of 2 levels of IDK. Note that $\Phi_2(\widehat{\Phi}(\mathcal{P})) = \widehat{\Phi}_2(\delta(g))$, where $g= \widehat{\Phi}(\mathcal{P})$. }
    \label{fig:idk2}
\end{figure}

Figure \ref{fig:idk2} demonstrates the process of 2 levels of IDK. A set of points generated from a distribution is mapped to a point in Hilbert Space $\mathscr{H}$ where $\Pi$ is the set of all mapped points. $\mathscr{H}_2$ is the second level Hilbert Space that is used to calculate Equation \ref{IDK2}.


Since IDK is also data-dependent, it enables an existing point anomaly detector to produce a high detection performance when the dataset has different distributions~\cite{ting2021isolation}.

\subsection{Applications}

\subsubsection{Trajectory data}

A trajectory can be defined as an ordered sequence of points.
We can use IDK to map each trajectory to a point $g$ in the Hilbert space, assuming that the points in a trajectory $X$ are i.i.d. from an unknown probability distribution $\mathcal{P}_{X}$ \cite{wang2024detecting}. Then we can obtain a new representation for each trajectory, i.e., $g=\widehat{\Phi}(\mathcal{P}_{X} | D)$. Thus, a set $D$ of $n$ trajectories gives rise to $\Pi= \{g_1,\dots,g_n\}$, and apply a point anomaly detector to compute the anomaly score for each $g \in \Pi$, corresponds to a trajectory in $D$.

\begin{figure}[!tb]
     \centering
     \begin{subfigure}[b]{0.23\textwidth}
         \centering
         \includegraphics[width=0.98\textwidth]{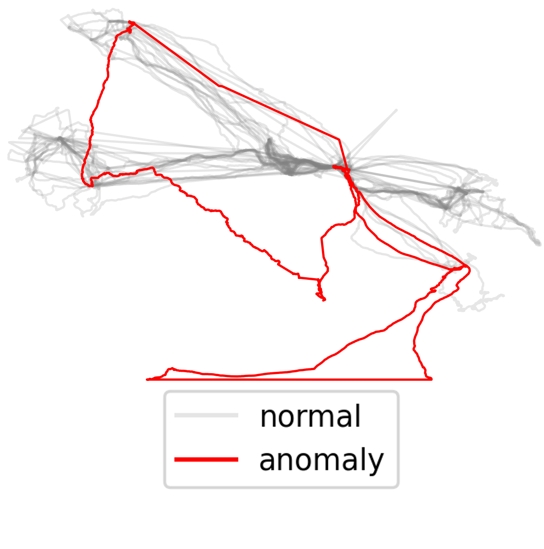}
         \caption{Anomalous trajectories}
     \end{subfigure}
     \begin{subfigure}[b]{0.23\textwidth}
         \centering
         \includegraphics[width=.98\textwidth]{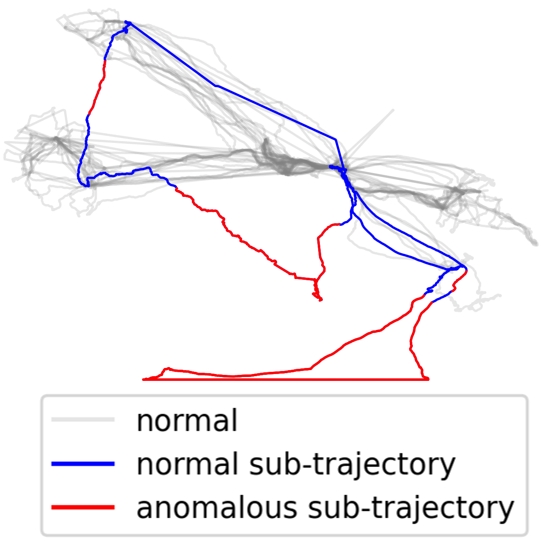}
         \caption{Anomalous  sub-trajectories}
     \end{subfigure}
        \caption{Detecting results of IDK on a trajectory dataset.}
        \label{fig:tra}
\end{figure}

With IDK as the anomaly detector, we can calculate the anomaly score of a trajectory $g$ using Equation \ref{IDK2} \cite{wang2024detecting}. IDK can also detect the abnormal sub-trajectories within an abnormal trajectory, by mapping each sub-trajectory to a point in Hilbert space and scoring it w.r.t. the average of kernel mean maps of all trajectories~\cite{wang2024detecting}. Figure \ref{fig:tra} shows the abnormal trajectories and sub-trajectories detected by IDK on a real-world dataset.

\subsubsection{Time series data}
When a time series $X$ is strictly stationary and has recurring sequences of length $m$, all sequences $X$ can be treated as sets of i.i.d. points generated from an unknown distribution in $\mathbb{R}$. Then IDK can be applied as a distributional treatment to detect anomalous subsequences $X_i$ in the series \cite{ting2022new,ting2024new}.

\begin{figure}[!tb]
    \centering
    \includegraphics[width=.49\textwidth]{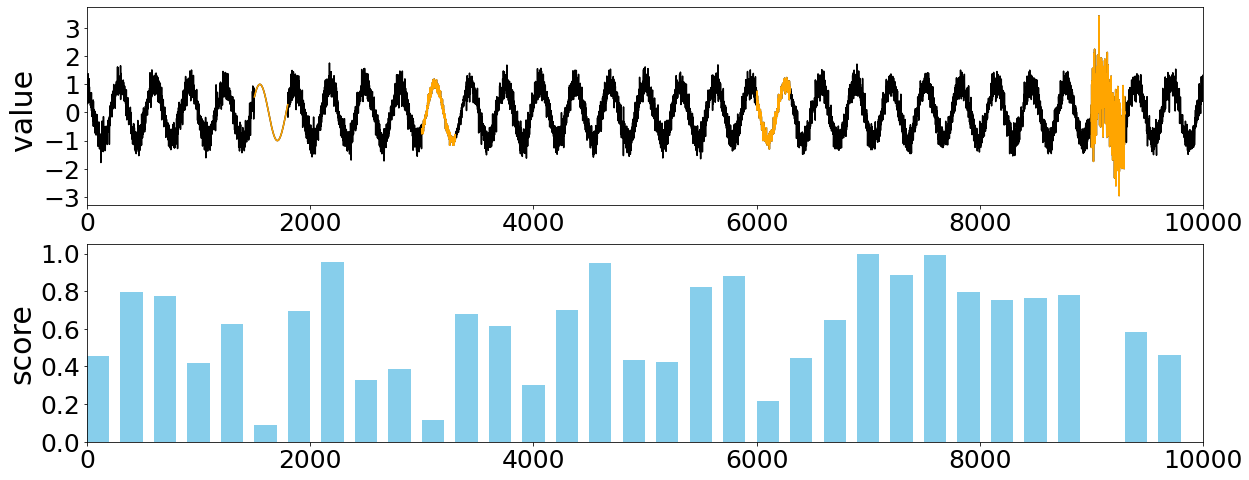}
    \caption{Illustration of $\widehat{\mathcal{K}}_2$ score for each subsequence in a time series.} 
    \label{fig:time}
\end{figure}

Similar to handling sub-trajectory, we can first map each subsequence in a time series to a point in Hilbert space based on the kernel mean map and score each subsequence w.r.t. the average of kernel mean maps of all subsequences. Figure \ref{fig:time} shows the $IDK^2$ score for each subsequence in a time series. The lower the score the more likely the subsequence would be anomalous.

Since this treatment does not involve point-to-point distance (e.g., \cite{yeh2016matrix,MPdist-2020,STOMP-2016,MPdist-2020,Shape-based-distance,paparrizos2019grail}) in the time domain or point-to-point alignment between two sequences (e.g., \cite{DTW-2005,DTW-SIAM2018,DTW-SIAM2019,DTW-BreakingQuadraticBarrier-2018}), IDK-based detector is the most effective with linear time complexity, compared with traditional detectors \cite{ting2022new,ting2024new}.

IDK is effective in identifying not only anomalous subsequences but also anomalous time series within datasets composed of non-stationary time series. We first encode the temporal order of each time series as an extra dimension using a $\mathbb{R}\times$time domain distributional treatment \cite{ting2024new}. This allows us to represent each time series as a point in Hilbert space. Then, we can apply a point anomaly detector or $\widehat{\mathcal{K}}_2$ to find anomalies in Hilbert space, which correspond to anomalous time series.



\section{Algorithms and demonstration}
Table \ref{table:categoriesofAD} lists the available source code of representative isolation-based anomaly detectors. A demonstration of the detection performance of 4 methods is shown in Table \ref{tab:demo}. The parameters of each algorithm are searched in a reasonable range. 
The results in Table \ref{tab:demo} show that iNNE can detect all anomalies on both datasets by adaptively partitioning the space and calculating relative anomaly scores. iForest fails to detect anomalies in datasets with irregular shapes and different densities. LSHiForest improves the detection on irregular shapes due to using non-axis parallel splitting, but still misses local anomalies.

~\label{sec:demo}
\begin{table}[!tb]
\caption{Summary of the representative work of isolation-based anomaly detection.}
\label{table:categoriesofAD}
\centering
\resizebox{\linewidth}{!}{
\begin{tabular}{l l}
\toprule

\textbf{Method} & \textbf{Code}\\
\midrule
iForest \cite{liu2008isolation} & \hyperlink{https://scikit-learn.org/stable/modules/generated/sklearn.ensemble.IsolationForest.html}{sklearn.ensemble.IsolationForest} \\

SCiForest  \cite{liu2010detecting} & \url{https://github.com/david-cortes/isotree} \\

EIF \cite{hariri2019extended}  & \url{https://github.com/sahandha/eif} \\

iNNE \cite{bandaragoda2018isolation} & \url{https://github.com/zhuye88/iNNE} \\

IK \cite{qin2019nearest, xu2021isolation} & \url{https://github.com/IsolationKernel/IK_Implementations} \\

LSHiForest \cite{zhang2017lshiforest} &  \url{https://github.com/xuyun-zhang/LSHiForest} \\

iForestASD~\cite{ding2013anomaly} & \hyperlink{https://pysad.readthedocs.io/en/latest/}{pysad.readthedocs.io} \\
HS-Trees~\cite{tan2011fast} & \hyperlink{https://pysad.readthedocs.io/en/latest/}{pysad.readthedocs.io}\\
RRCF~\cite{guha2016robust}  & \hyperlink{https://pysad.readthedocs.io/en/latest/}{pysad.readthedocs.io}\\

IDK-Time Series~\cite{liu2012isolation} & \url{https://github.com/IsolationKernel/TS} \\

IDK-Trajectory~\cite{wang2024detecting} & \url{https://github.com/IsolationKernel/TrajectoryDataMining} \\

DeepiForest \cite{xiang2023deep} & \url{https://github.com/xiagll/DeepiForest} \\

DIF \cite{xu2023deep} & \url{https://github.com/GuansongPang/deep-iforest} \\

OptIForest \cite{xiang2023optiforest} & \url{https://github.com/xiagll/OptIForest}\\

DOIForest \cite{xiang2023deep} & \url{https://github.com/xiagll/DOIForest} \\

\bottomrule
\end{tabular}}
\end{table}

\begin{table}[!hbt]
    \centering
    \caption{Demonstration of the best performance of 3 methods on two synthetic datasets in terms of AUC-ROC (Receiver Operating Characteristic)~\cite{provost1998case} and AUC-PR (Precision-Recall)~\cite{manning1999foundations}}
    \label{tab:demo}
    \begin{tabular}{cll}
    \toprule
        Method  &   {Dataset 1}   &   {Dataset 2}  \\
    \midrule
        \begin{turn}{90}    iForest \end{turn}  & \includegraphics[width=1.2 in]{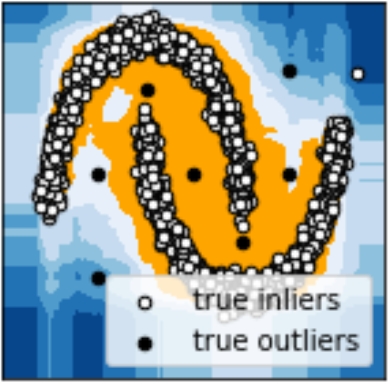} &\includegraphics[width=1.2 in]{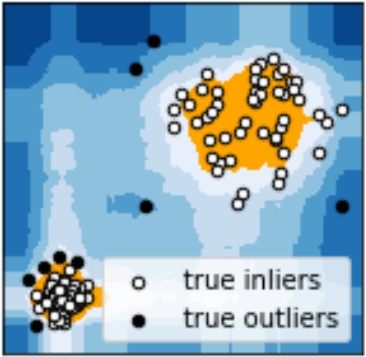}  \\
        & AUC-ROC $= 0.98 $ & AUC-ROC $ = 0.99$\\
        & AUC-PR $ = 0.46  $ & AUC-PR $ = 0.82 $\\ \midrule
        \begin{turn}{90}   LSHiForest \end{turn} & \includegraphics[width=1.2 in]{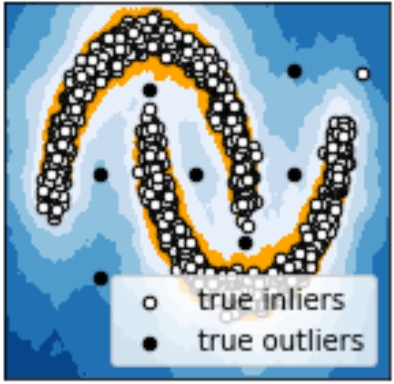} & \includegraphics[width=1.2 in]{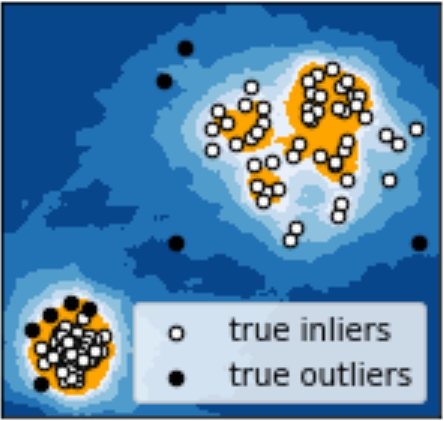}   \\
        & AUC-ROC $= 0.998$ & AUC-ROC $ = 0.95 $\\
        & AUC-PR $ = 0.84 $ & AUC-PR $ = 0.66 $\\ \midrule
        \begin{turn}{90}  iNNE \end{turn}  & \includegraphics[width=1.2 in]{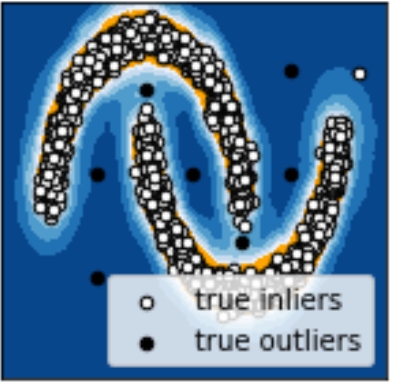} &\includegraphics[width=1.2 in]{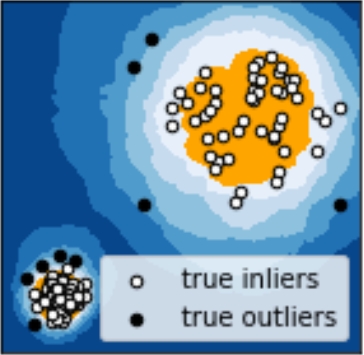}     \\
          & AUC-ROC $= 1$ & AUC-ROC $ = 1$\\
        & AUC-PR $ = 1$ & AUC-PR $ =  1$\\
    \bottomrule
    \end{tabular}
\end{table}

\section{Parameter settings and model optimization}

Isolation-based anomaly detection algorithms have two key parameters: the sub-sampling size $\psi$ and the number of ensembles $t$. The original paper of iForest recommends to use $\psi=256$ and $t = 100$ as the default. However, the parameters should be tuned based on a given dataset, so we consider the range of  $\psi \in [2^1, 2^2, \dots, 2^{10}]$. We can fix $t$ to 100 or 200, as a larger value of $t$ tends to produce more stable results but also longer running time. 

It would be unfair to compare the performance of isolation-based methods with fixed parameter settings against other methods, e.g., \cite{paulheim2015decomposition,ouardini2019towards,hayashi2021less}, that have undergone rigorous parameter tuning to find the best configurations. Furthermore, applying isolation-based techniques directly to unstructured data may not be appropriate, as these methods are typically designed to handle structured and numerical data.


Based on some heuristic methods, we can conduct parameter and model optimization as follows:

\subsection{Parameter optimization}
Since anomalies are assumed to be rare, a few instances will have significantly higher scores while the majority of instances will keep their scores at a minimum level, i.e., the anomaly scores of most points should be similar. 
Therefore, we may search for the best parameter $\psi$ in IK-based anomaly detection that leads to maximum stability of anomaly score distribution, i.e.,

\begin{equation}
    \psi^* = \mathop{\mathrm{argmin}}\limits_{\psi}{\Bar{E}(C_{\psi})},
        \label{eqn:psi}
\end{equation}

\noindent where $C_{\psi}$ are the anomaly scores generated by different $\psi$; $\Bar{E}(.)$ is a measure of instability, e.g., variance or Gini coefficient~\cite{gini1912variabilita}. Note that a lower $\Bar{E}(.)$ value means a higher stability. This search method performs well in changing point detection using IDK on streaming datasets~\cite{cao2024detecting}.

\subsection{Model optimization}
Given a subsample size, the width and the depth of the tree compete with each other, When building multi-fork isolation trees, OptIForest \cite{xiang2023optiforest} proposes an objective function to optimize the isolation efficiency of LSHiForest \cite{zhang2017lshiforest}i.e., maximize the number of leaf nodes within a given constant isolation area wrt the branching factor, to maximise the differentiating/isolation ability of the trees.\footnote{Inspired by the Huffman coding~\cite{huffman1952method}, isolation area is the product of tree depth and average branching factor, to measure the cost of encoding. The branching factor is the number of branches for each node. }

It gives the first theoretical explanation for the isolation forest mechanism. It also proves that the optimal number of branches is Euler’s number $e$. Based on this finding, OptIForest applies clustering-based learning to hash for a trade-off between bias and variance for data isolation.

To lower the optimization cost of OptIForest, DOIForest \cite{xiang2023deep} employs the genetic algorithm to learn the optimal isolation forest and fine-tune the parameters in data partitioning. DOIForest uses two mutation schemes: (1) randomly re-constructing an entire iTree with new $\psi$ subsampling points; and (2) re-assigning points to the hash bins with a new offset $b$ value in a randomly chosen node of an LSHiTree.

\section{Discussions and prospect}
\label{sec:discussion}
 

\subsection{Isolation mechanisms applied to tasks not related to anomaly detection}
Since the concept of isolation was introduced, it has not been confined to point and group anomaly detection only. The isolation mechanism has been applied to  numerous other tasks, including:
\begin{itemize}
    \item Clustering: a new class of clustering algorithm based on point-set Kernel can cluster a massive number of points in a few minutes linear time complexity, e.g., psKC~\cite{ting2022point}, StreaKHC~\cite{han2022streaming}, TIDKC~\cite{wang2023distribution} and IDKC~\cite{zhu2023kernel}. 
    \item Emerging new class detection: a task which is seemingly unrelated to anomaly detection, but has been shown that it can be addressed effectively using an isolation-based anomaly detector, i.e., emerging new class~\cite{mu2017classification, cai2019nearest}. 
    \item Classification: multi-instance learning~\cite{xu2019isolation} and graph classification via isolation Graph Kernel~\cite{xu2021isolation}.
    \item Kernel estimators and Regression: Ting et al.~\cite{ting2021isolation1, ting2023isolation} proposed a novel approach to density estimation and regression using isolation kernels, which enables adaptive kernel density estimators (IKDE) and kernel regressions (IKR) with constant-time complexity for each estimation. This means that the resulting estimators have the same computational complexity as the original isolation kernel, making them the first KDE and KR that are both fast and adaptive.
    \item Change-point/interval detection: isolation-based change interval detector (iCID)~\cite{cao2024detecting} can identify different types of change-points in streaming data with the tolerance of outliers and linear time complexity.
    \item Topological data analysis: a newly proposed  kernel-based filter function ~\cite{zhang2023towards}, which is inspired by Isolation Kernel, is robustness against noise and varying densities within a point cloud for topological data analysis.
\end{itemize}

\subsection{Future directions}~\label{sec:conclusion}

There are some open challenges and future directions for isolation-based methods as follows.

\subsubsection{Theoretical analysis and probabilistic explanation} There are few theoretical analyses of the properties of different isolation mechanisms. How to directly calculate or statistically infer the anomaly scoring without conducting sampling on the data is a promising but underexplored research direction, which could improve detection, scalability, and responsiveness.

\subsubsection{Incremental learning} For streaming data, most isolation-based methods rebuild the model from scratch when new data comes, which is inefficient and impractical for large or fast data streams. Incremental learning techniques are needed to update the models without losing accuracy or adding much computation.

\subsubsection{Partitioning optimization} Partitioning strategies have a significant impact on the performance of anomaly detection based on isolation mechanisms. Current optimization techniques are mainly for tree-based models. Additionally, most partitioning methods are unsupervised, ignoring any prior knowledge or expert feedback. It is worth exploring how to optimize the splitting method using limited labeled data to achieve better detection.


\begin{acks}
We appreciate the suggestions from Shuaibin Song, Zijing Wang, Zongyou Liu from Nanjing University and Dr Xuyun Zhang from Macquarie University. 

This project is supported by the National Natural Science Foundation of China (Grant No. 62076120).

\end{acks}



\clearpage
\newpage

\bibliographystyle{ACM-Reference-Format}
\bibliography{ref}



\end{document}